# Video-based Cross-modal Auxiliary Network for Multimodal Sentiment Analysis

Rongfei Chen, Wenju Zhou, Yang Li, Huiyu Zhou

***Abstract*—Multimodal sentiment analysis has a wide range of applications due to its information complementarity in multimodal interactions. Previous works focus more on investigating efficient joint representations, but they rarely consider the insufficient unimodal features extraction and data redundancy of multimodal fusion. In this paper, a Video-based Cross-modal Auxiliary Network (VCAN) is proposed, which is comprised of an audio features map module and a cross-modal selection module. The first module is designed to substantially increase feature diversity in audio feature extraction, aiming to improve classification accuracy by providing more comprehensive acoustic representations. To empower the model to handle redundant visual features, the second module is addressed to efficiently filter the redundant visual frames during integrating audiovisual data. Moreover, a classifier group consisting of several image classification networks is introduced to predict sentiment polarities and emotion categories. Extensive experimental results on RAVDESS, CMU-MOSI, and CMU-MOSEI benchmarks indicate that VCAN is significantly superior to the state-of-the-art methods for improving the classification accuracy of multimodal sentiment analysis.**

## I. INTRODUCTION

MULTIMODAL sentiment analysis has been widely applied in real life, such as emotion-based content recommendation [1], semantic multimedia indexing and retrieval [2], and harmful horror video detection [3]. However, it is still a challenge to analyze sentiment more effectively and accurately. One of the main challenges of multimodal sentiment analysis is the increased prediction error caused by insufficient unimodal representation. Specifically, unimodal representations can only describe changes in emotion from a single perspective [4]. Therefore, compared with modeling unimodal information, previous research has focused on the use of specific deep neural networks (DNNs) to efficiently learn the joint representation of multiple modalities [5]. For instance, a large number of studies seek to tackle these challenges by building complex network structures [6] and fusing multimodal feature matrices [7], which can mine deep multimodal features and enhance interaction between audiovisual signals, respectively. However, the introduction of complex network structures inevitably reduces efficiency and generates data redundancy. These redundant features are generally derived from two aspects: Firstly, invalid audiovisual fragments such as tone words and repeat frames are inevitably introduced during fusing multimodal data [8]. Secondly, the weakly correlated features are redundant as they contribute less to improving accuracy, which should be filtered. The redundant features not only increase the computational effort but also interfere with the optimization. Therefore, a series of algorithms have been proposed recently to eliminate the above disadvantages. The conventional method in sentiment analysis is Convolutional Neural Network (CNN) [9]. In CNN-based approaches, the visual features are extracted through convolution and pooling operations in CNN, and then the classification results are output according to these visual features [10]. The disadvantage of the CNN-based model is that all input images are processed without filtering the weakly correlated video frames. In other words, this model assigns the same weight to each frame in the video without highlighting keyframes. Therefore, enhancing the effective sentiment representation of the single modality and eliminating redundant features are key to improving the efficiency and accuracy of multimodal models.

In this paper, a Video-based Cross-modal Auxiliary Network (VCAN) is proposed to extend the multi-scale acoustic representations and reduce redundant computation. The VCAN consists of the Audio Features Map Module (AFMM) and the Cross-Model Selection Module (CMSM). The AFMM is implemented by combining the EMD-based speech signal decomposition and the K-means clustering. And the CMSM is designed to select keyframes from video with the aid of audio modality, which aims to reduce the redundancy of video frames in data fusion. The contributions of our research are highlighted as follows:

- A novel bimodal interaction model, i.e., the video-based cross-modal auxiliary network (VCAN), is proposed for multimodal sentiment analysis. This model can simplify the multimodal sentiment analysis to an image classification, which significantly improves the accuracy of multimodal sentiment analysis.
- An acoustic feature extraction module (AFMM) is presented to enhance the multi-scale acoustic sentiment representations by combining signal decomposition (EMD) and feature clustering (K-means). Moreover, a Cross-Model Selection Module (CMSM) is proposed, which not only improves the

This work was supported by the National Natural Science Foundation of China (No. 61877065). *Corresponding author*: Wenju Zhou.

Rongfei Chen, Wenju Zhou and Yang Li are now with the School of Mechatronic Engineering and Automation, Shanghai University, Shanghai 200444, China (e-mail: chenrongfei@shu.edu.cn, zhouwenju@shu.edu.cn, dreyang@163.com). Huiyu Zhou is with School of Computing and Mathematical Sciences, University of Leicester, Leicester, LE1 7RH, United Kingdom (e-mail: hz143@leicester.ac.uk).

Color versions of one or more of the figures in this article are available online at http://ieeexplore.ieee.org.

interactivity of audiovisual modalities through a cross-mode selection mechanism but also eliminates redundant computations by extracting video keyframes.
- The experimental results demonstrate the validity of our model when we compare its result with the tri-modal algorithms and bimodal methods on the RAVDESS, CMU-MOSI, and CMU-MOSEI datasets. Moreover, we briefly investigate the asymmetric contribution of each mode in the joint representation and the feasibility of combining language modality with VCAN.

The remainder of this paper is organized as follows. Section II gives an overview of related work. Section III describes our proposed method. Section IV presents the implementation details and experimental results on the public multimodal sentiment analysis datasets. Finally, we conclude with the major contributions of this work in Section V.

## II. Related Work

VCAN consists of three components: acoustic feature extraction, visual feature extraction, and audiovisual modalities fusion. The related work of these three components is described below.

### *A. Acoustic features extraction*

Extracting the emotional content (speech features) from a speech signal and identifying the sentiment polarities is an important task for researchers. Currently, the speech features used for sentiment analysis can be categorized into prosodic features, sound quality features, spectrograms, etc. [11]. For example, the duration, pitch, energy, zero-cross rate, and formant are typical prosodic and sound quality features, which are the low-level descriptors of audio signals [12]. The spectrogram-based method attracts more attention in Speech Emotion Recognition (SER). The spectrogram makes it possible to migrate high-performance CNN models to acoustic spectrogram-based speech emotion recognition because spectrograms can convert 1D sequences into 2D images [10]. For example, *Shehu et al*. [13] proposed an enhancement approach to remove the limitations of overfitting on small training datasets, which employed a multi-windows enhancement strategy to increase the representative number of spectrograms. *Satt et. al.* [14] presented a new implementation of speech emotion recognition that directly used spectrograms for a deep neural network to improve the classification accuracy. *Turgut* [15] utilized four different texture analysis methods to investigate the effect of spectrogram images on speech emotion recognition. These methods have obtained promising results in spectrogram-based speech emotion recognition, but most of them only explore the image characteristics of spectrograms without considering their acoustic features.

Additionally, the shortage of available speech samples also affects the accuracy of SER [13]. Some audio signals with strong noise are difficult to directly analyze the emotional components. Thus, many scholars leveraged signal decomposition to address the aforementioned problem. *Leila et. al.* [16] used the combination of empirical mode decomposition (EMD) and the Teager-Kaiser Energy Operator (TKEO) to extract acoustic features. Similarly, *Pan et al.* [17] proposed a strategy for SER by combining the Evolutional Algorithm (EA) with the Empirical Mode Decomposition (EMD) to improve the emotion recognition rate. *Hou et al.* [18] investigated the function of multi-view speech spectrograms, which includes extracting multi-view features by the attention network and the collective relation network. The results of the above methods show that it is feasible to exploit the multi-view speech representations. However, increasing the sample diversity of audio sequences inevitably generates weakly correlated features, which result in data redundancy [19]. So, it is necessary to streamline the expanded data using clustering or other data processing methods.

The AFMM is proposed to enhance the diversity of acoustic representation. It is implemented by combining the EMD-based speech signal decomposition and the K-means clustering. Compared to the aforementioned models, the AFMM increases the multi-scale emotional representation of speech by decomposing the speech signal into sub-sequences and extracting the acoustic features of each sub-sequence. In addition, AFMM introduces unsupervised clustering methods to achieve automatic clustering of acoustic features, which eliminates feature redundancy and information loss due to the manual grouping of parameters. See *III.D. Feasibility Analysis of Clustering Methods in AFMM* for detailed analysis and discussion.

### *B. Visual features extraction*

The visual modality (i.e. facial images) can provide a rich source of expression features for judging emotion. Since the images contain more general emotional information than voice sequences, sentiment analysis with the help of facial images has been widely researched in recent years. Typically, traditional facial emotion recognition (FER) focuses on extracting some manual features such as facial texture, facial contour, and organic relative positions, which estimate individual affective states by measuring feature changes over a continuous period [20]. However, these fundamental features are inadequate for recognizing emotion in a complex environment, such as identifying emotions in low-light outdoor conditions.

Therefore, using deep learning neural networks for facial feature extraction and classification becomes popular in recent years [21]. For example, *M. Kalpana et al.* [22] transferred the DNN including Resnet50, VGG-19, Inception-V3, and Mobile-Net to recognize facial emotion, which validated the effectiveness of DNN on FER. Considering the temporal properties of emotion changes, some DNNs with contextual information are widely used. *Huan et al*. [23] proposed a video multimodal emotion recognition method based on Bi-GRU and attention fusion to improve the accuracy of emotion recognition in time contexts. *Zhang et al.* [24] considered the real-time requirement of the model and proposed a novel multimodal emotion recognition model for conversational videos based on reinforcement learning with domain knowledge, and this paper achieved state-of-the-art results on a weighted average. However, as the consecutive images selected from visual fragments are highly similar to each other, it is time-consuming to extract valid features from these similar frames. Therefore, selecting video keyframes to eliminate redundant frames may contribute to improving the efficiency of feature extraction.

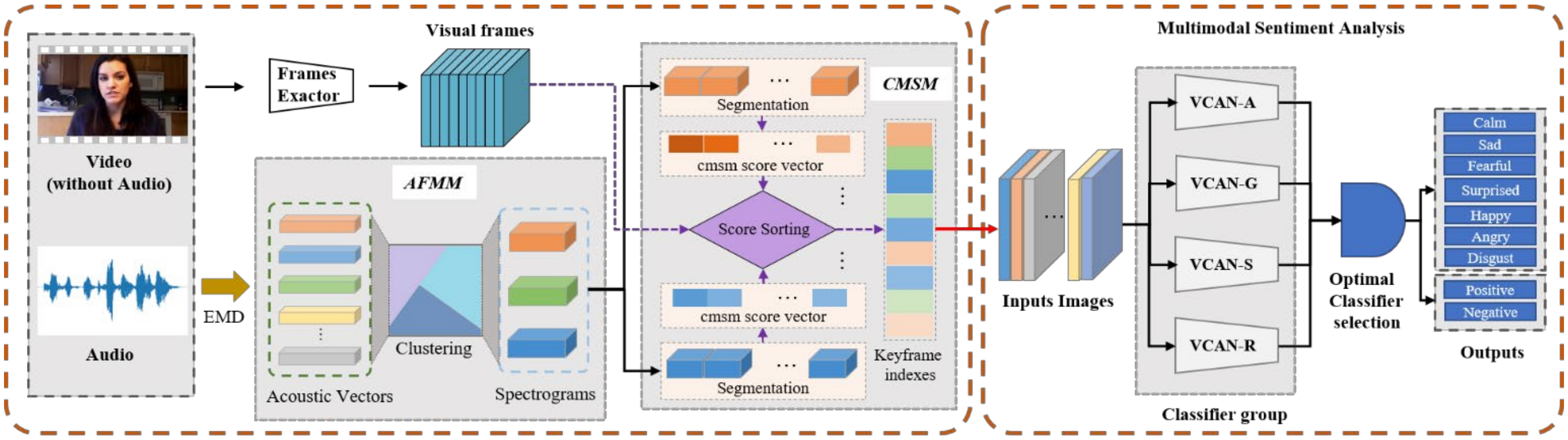

**Fig. 1.** The overall framework of the Video-based Cross-modal Auxiliary Network (VCAN).

The CMSM is essentially a cross-modal video keyframe selection method, which generates a video keyframe index by calculating the mathematical statistics of the related spectrogram. Unlike other methods that fuse audiovisual feature vectors into a matrix of affective representations, CMSM can automatically select video keyframes by recognizing the emotional changes of speech modality based on the semantic and temporal synchronization characteristics of audiovisual modality. See *IV.D Parametric experiments and analysis (Video Keyframes Selection and Feasibility Study of Auxiliary Mechanisms)* for detailed analysis and discussion.

### *C. Audiovisual information fusion*

Since multimodal analysis can leverage both independent and complementary information to provide comprehensive representations [25], such a technique has drawn much interest in sentiment analysis [26]. The multimodal fusion strategy can be grouped into feature-level fusion [27] and decision-level fusion, and the latter is the current mainstream fusion approach [28]. For example, *Hossain et al.* [29] proposed a model that integrated the CNN-based feature extraction, ELMs-based features fusion, and SVM-based decision outputs. Similarly, *Schoneveld et al.* [30] presented a deep-learning approach for audio-visual emotion recognition, which recently leveraged advanced algorithms such as knowledge distillation and deep architectures. *Nemati et al.* [31] investigated the advantages of latent space linear maps and designed a hybrid multimodal data fusion model for audiovisual features fusion.

The video-based audiovisual fusion focuses more on capturing the complex spatiotemporal and semantic relationships among consecutive video frames. Under this condition, the Recurrent network with its variants and the attention mechanism [32] are widely used in multimodal fusion [26]. For example, *Ou et al.* [4] extended the attention mechanism to obtain a global representation of effective video. These global representations are selected from the local features obtained by the attention mechanism. Although existing audiovisual fusion models are capable of obtaining effective joint representations, they are more complex and difficult to explain [5]. As described in [4], selecting key local components from the global representation is beneficial for reducing the complexity of the model. Thus, selecting keyframes from the video may be an effective method for improving model performance. Furthermore, various video-based sentiment analysis tasks derived from multimodal information fusion such as drowsiness recognition [33], facial performance editing [34], crowd behavior analysis [35], and human action expression recognition [36] have received extensive attention.

## III. Methodology

As shown in Fig.1, our model contains two independent inputs, i.e. the audio input and the video input. They are part of the same audio-video sequence. First, the AFMM is used to encode features from the audio modality, where the acoustic vectors and three spectrograms are generated through EMD decomposition and K-means clustering, respectively. Then, the CMSM is employed to obtain the key-frame indexes. Specifically, the score vectors are calculated based on segmented spectrograms, and the key-frame indexes are selected by integrating acoustic score vectors and visual frames. Next, a collection of keyframes according to key-frame indexes are fed into the classifier groups. Finally, the optimal outputs of the classifier group are used for the Multimodal Sentiment Analysis (MSA) to determine the polarity of emotions (i.e. Positive and Negative) and the category of emotion (i.e. Calm, Happy, Sad, Angry, Fearful, Disgust, Surprised, etc.), respectively. The detailed analysis of each module is described below.

### *A. The Acoustic Feature Mapping Module (AFMM)*

The AFMM is proposed as the audio signal processing component of the VCAN. The main function of this module is to balance the relationship between sample diversity and data redundancy in audio. In AFMM, the EMD algorithm expands the multi-scale representation of acoustic features by decomposing the audio signal, and the K-means algorithm is utilized to reduce data redundancy by reconstructing subsequences with high similarity [19].

Empirical Mode Decomposition (EMD) is a method for dealing with non-stationary signals, which decomposes the signal according to the time scale characteristics of the data without presetting any basis functions [16]. The essence of the EMD algorithm is to identify and extract all Intrinsic Mode Functions (IMFs) contained in the signal through the characteristic time scale. The advantage of the EMD method over other signal decomposition methods is that it can intuitively extract the main components of different frequency bands and adaptively represent the local characteristics of the signal [17]. Initially, EMD algorithms were widely used in signal processing fields such as mechanical fault diagnosis. In recent years, EMD has been applied to the analysis and enhancement of acoustic features [18].

The role of the EMD algorithm in AFMM is to increase the multi-scale emotional representation of speech by decomposing the audio signal. Therefore, given an input audio set is $\mu(t) = \left(\mu_1(t), \mu_2(t), \dots, \mu_a(t)\right)$, the process of decomposing an original signal into several subsequences utilizing the EMD algorithm can be summarized as follows:

$$\left(Imf_a^S(t), Imf_{res}(t)\right) = \mathrm{E}(\mu_a(t), \theta) \tag{1}$$

where $\mathrm{E}(\cdot)$ is the EMD function that includes two input variables: $\mu_a(t)$ denote the *a-th* raw audio sequence and θ is a collection of EMD's parameters. The set θ contains two input parameters, i.e. $\theta = \{\mathcal{S}, \mathcal{U}\}$, where $\mathcal{S}$ denotes the threshold at which the overall sifting process will stop (default value: 1e-8) and $\mathcal{U}$ represents the maximum number of IMFs to compute. Each $Imf_a^S(t)$ represents the *S-th* intrinsic mode functions of $\mu_a(t)$, which can reconstruct the raw signal by combining it with the remainder item $imf_{res}(t)$, i.e., $\mu_a(t) = \sum_{i=1}^{S} Imf_a^i(t) + Imf_{res}(t)$

$$\text{s.t.} \begin{cases} \sum_{t=0}^{T} \dfrac{\left|Imf_a^{S-1}(t) - Imf_a^S(t)\right|}{\left(Imf_a^{S-1}(t)\right)^2} < \varepsilon \\ S \leq \epsilon \end{cases} \tag{2}$$

where $T$ denotes the total number of samples of the input signal, which is a dynamic variable determined by the duration of the input samples. Equation (2) is the constraints of decomposition that ensure the validity of the intrinsic mode function, where ε and ϵ denote the decomposition stop threshold and the decomposition maximum layer threshold, respectively.

The EMD algorithm focuses on the evolution of the overall signal by fitting the upper and lower envelopes of audio curves. This principle is consistent with personal emotional development, which emphasizes changes in the overall state rather than ignoring the cumulative nature of emotions over time. Furthermore, the frequency characteristics of speech play an important role in identifying the type of emotion. For example, the parameters of the formant, including formant frequency, bandwidth, and amplitude, are crucial features for analyzing the emotional state of speech. A more detailed and relevant analysis can be found in [37]. The reason for combining audio signal decomposition with frequency-based acoustic feature extraction is manifested in two aspects, which are, expanding the quantitative representation of emotions in different frequency components and fitting the formant of speech from another perspective. However, due to the frequency aliasing, it is complicated to analyze the emotion-related acoustic representations. The signal decomposition method can eliminate the frequency aliasing by decomposing the raw sequence into several distinct sub-sequences. Therefore, the EMD decomposition algorithm is introduced to increase the emotional representation of the audio in different frequencies.

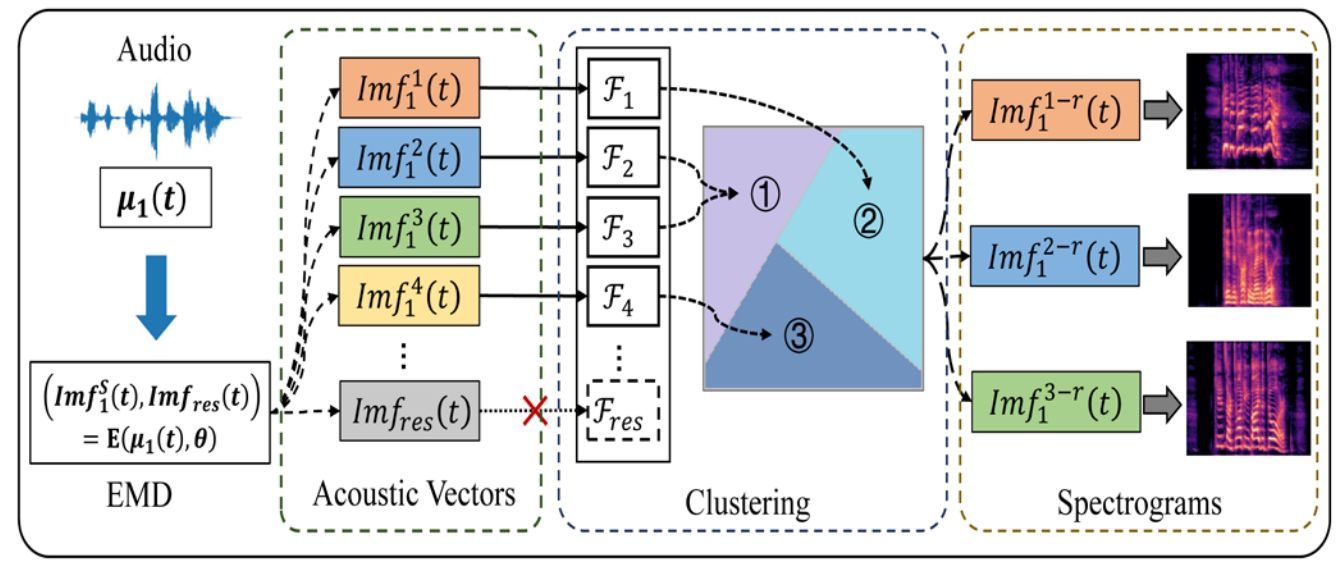

**Fig.2.** The schematic of the Acoustic Feature Mapping Module (AFFM)

A detailed discussion will be given in *Section IV-D*. The formant information is contained in the envelope of the speech spectrum. Coincidentally, obtaining the temporal envelope of the signal is a necessary and sufficient condition for the EMD algorithm. Therefore, we can indirectly get formant information via obtaining the signal temporal envelope. The connection between the formant and the envelope will be further explained in *Section III-C*.

The schematic of the Acoustic Feature Mapping Module (AFFM) is presented in Fig.2. The function of AFMM is to decompose the speech signal into subsequences of different frequency bands and convert them into spectrograms. As shown in Fig.2, the audio $\mu_1(t)$ is randomly selected as the input sequence to AFMM. According to the EMD calculation formula, the input sequence $\mu_1(t)$ is first decomposed into different acoustic vectors (or named Intrinsic Mode Functions, IMFs, i.e., $Imf_1^1(t)$, $Imf_1^2(t)$, $Imf_1^3(t)$, $Imf_1^4(t)$, …, $Imf_{res}(t)$). The related acoustic features such as $\mathcal{F}_1$, $\mathcal{F}_2$, $\mathcal{F}_3$, and $\mathcal{F}_4$ are extracted from these IMFs. The solid arrows '→' in the diagram represent the acoustic feature extraction process. Notably, since the remainder item $imf_{res}(t)$ contains a lot of noise and interferes with the subsequent clustering results, the features of $imf_{res}(t)$ are filtered out. Then acoustic features are classified into different clusters by k-means clustering. The reconstructed acoustic vectors $Imf_1^{1-r}(t)$, $Imf_1^{2-r}(t)$, $Imf_1^{3-r}(t)$ are generated by superimposing the IMFs in the same cluster. Finally, the reconstructed acoustic vectors are converted into the MEL spectrograms.

The spectrogram is a graph showing the change in speech spectrogram with time, which shows a large amount of information related to the utterance characteristics of speech. Since the spectrogram integrates the three attributes (i.e. temporal character, frequency characters, and power) of the speech in a visual modality, the spectrogram plays an important role in describing the emotional dynamics and sentiment analysis. To obtain acoustic features that match the auditory characteristics of the human ear, the speech spectrogram is often transformed into a MEL spectrogram by using MEL-scale filter banks. Using MEL spectrograms as the input of audio modality is a popular audio processing method. And some MEL spectrogram-based models have achieved good classification results in Speech Emotion Recognition (SER).

The classification accuracy benefits from the sample diversity of the signal decomposition, but the data redundancy induced by the similarity between subsequences cannot be ignored [38]. Moreover, signal decomposition inevitably results in information loss. To cope with the data redundancy and the information loss, the K-means method is introduced as a data dimensional reduction function of the AFMM. Furthermore, we reproduce parameter settings such as optimal values and the number of clusters from [38] to demonstrate the advantages of unsupervised clustering methods compared to manual parameter settings. Experimental results and a detailed discussion are given in Section III-*D*.

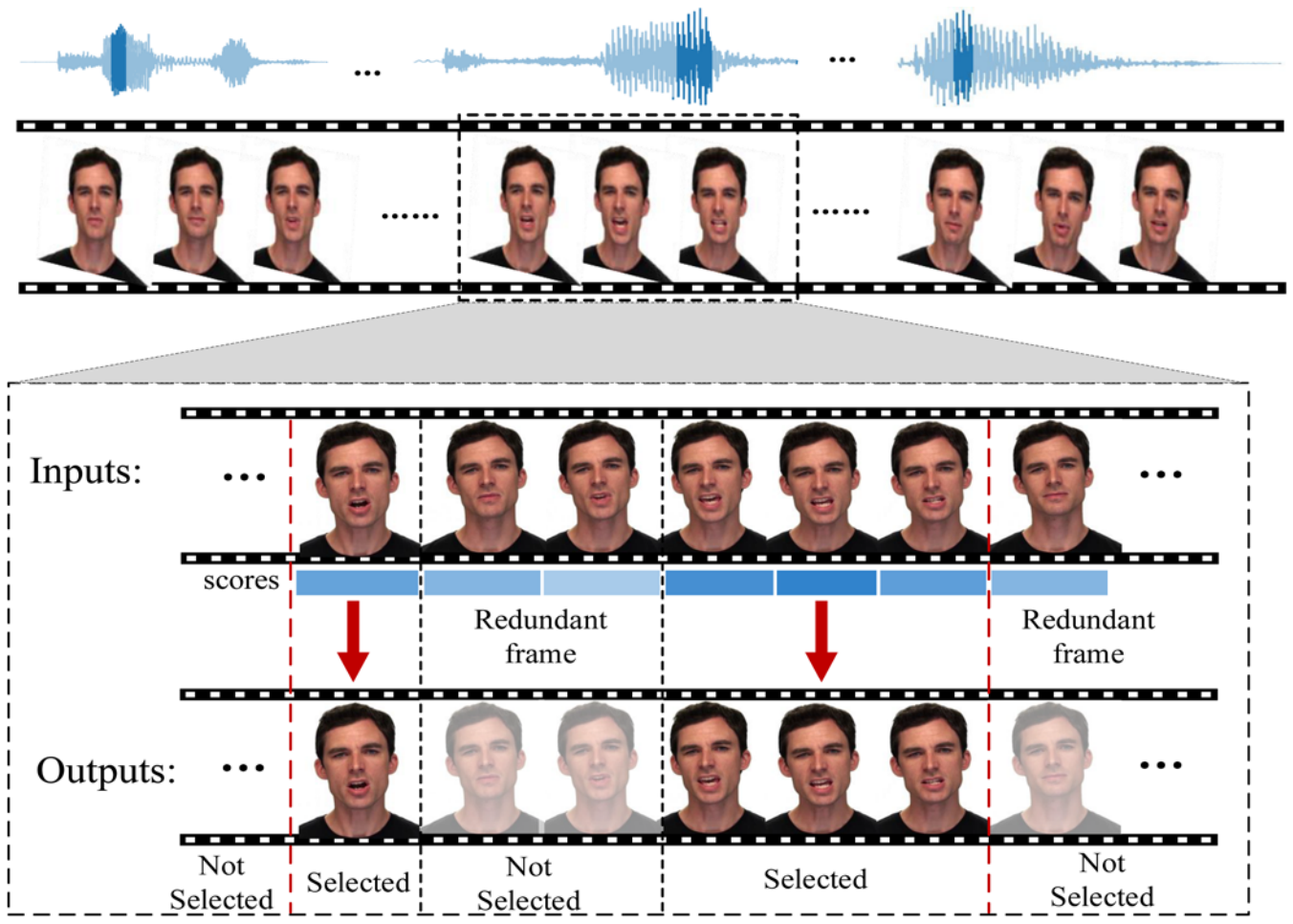

**Fig. 3.** The principle of CMSM.

### *B. The Cross-Modal Selection Module (CMSM)*

We adopted the expression of k-Neighborhood Dispersion from [37] and adjust its parameters for measuring emotional fluctuations. Then the Cross-Modal Selection Mechanism (CMSM) is built to explore cross-modal effective interactions and reduce frame redundancy. The principle of CMSM is shown in Fig.3, where the redundant data includes frames corresponding to both low CMSM scores and invalid fragments.

Unlike the traditional strategy that utilizes complex mathematical matrices to map multimodal information into a common subspace, CMSM uses auxiliary modality (i.e. audio modality) to guide the video to extract affective keyframes. The principle of CMSM is to use the audio features with drastic emotion fluctuations to match the representative images of the video. So, it can transform complex video-based sentiment analysis into a straightforward image multi-classification task. Formally, given the input videos are $\mathcal{V}(t) = (\mathcal{V}_1(t), \mathcal{V}_2(t), \ldots, \mathcal{V}_a(t))$, $\mathcal{V}(t)$ consists of multiple consecutive video frames. For example, $\mathcal{V}_1(t)$ represents the first frame of $\mathcal{V}(t)$. The cross-modal selection mechanism can be expressed as:

$$\chi^m = \mathcal{F}\left(\bigcup(\mathcal{M}_j^n)\right) = \mathcal{F}\left(\bigcup(\varphi \odot \gamma_j^n)\right), \quad j = 1,2,\ldots,\mathcal{J}.\ n = 1,2,\ldots,\mathcal{N} \tag{3}$$

where $\chi^m$ denotes the *m-th* key-frame set gained by the judgment condition $\mathcal{F}(\cdot)$, which ensures that the sequence value is unique. The symbol $\bigcup(\cdot)$ denotes the union of a set. $\mathcal{M}_j^n \in \mathbb{R}^{\mathcal{N}\times\mathcal{J}}$ is a MEL-matrix of $Imf_a(t)$ that is obtained using MEL filters, where $\mathcal{N}$ (default value: $\mathcal{N} = 26$ ) and $\mathcal{J}$ (default value: $\mathcal{J} = 77988$) denote the number of MEL filters and duration of the video (i.e. the number of frames), respectively. $\varphi: \mu_a(t) \rightarrow \mathcal{V}_a(t)$ is the static mapping connection that maps audio frames $\mu_a(t)$ to video frames $\mathcal{V}_a(t)$. The mapping relationship can be roughly understood as the synchronization between sound and picture in a video, which is determined by the frame rate of the dataset. $\gamma_j^n$ denotes a measure of sentiment fluctuations, which can calculate the mathematical statistics per frame:

$$\gamma_j^n = \nu_j^{n\left(\alpha_j^n + \beta_j^n\right)} = \left(\nu_j^n\right)^{\alpha_j^n} \cdot \left(\nu_j^n\right)^{\beta_j^n} \tag{4}$$

where $\nu_j^n$ is variance, which is used to distinguish valid speech clips from invalid ones. Both $\alpha_j^n$ and $\beta_j^n$ are calculated from the characteristics of the spectrogram at a fixed period. The $\alpha_j^n$ is the quotient of the mean value of the pitches in the real voice frequency band (i.e. 64Hz-1.1kHz) and the sum of the pitches in all frequency bands. Differently, the $\beta_j^n$ replaces the mean value presented in $\alpha_j^n$ with the maximum value of the pitch in the full frequency band. Both of these parameters con MC and RW formant information in the acoustic spectrogram, which aims to improve emotional expression by utilizing the local short-term information of the formant.

### *C. Time-frequency Variable Conversion Analysis*

As described in CMSM, the spectrogram is a link between acoustic features (i.e. formants and pitches) and visual images (i.e. video key-frames) to enable the information transition in the video. Its properties have different functions in CMSM: the temporal property focuses on solving the subsequent alignment of the audiovisual signal, i.e. time alignment of video keyframes to speech segments; the frequency attribute is employed to determine the available frequency threshold, i.e., the frequency range of the signal should match the normal vocal frequency range (male voice: 64Hz-698Hz. female voice: 82Hz-1.1kHz); and the amplitude property is utilized to depict formant information in the subsequent calculations.

To investigate the frequency domain properties of formants in different subsequences, a Fourier transform example is introduced to demonstrate the relationship between the frequency formants and the temporal envelope. We assume that the signal $f(t)$ is a signal with period T consisting of a constant and several trigonometric functions, and its Fourier series is expressed as follows:

$$f(t) = \alpha_0 + \sum_{n=1}^{\infty} \alpha_n \cdot \cos(n\omega t + \varphi) \tag{5}$$

where $\alpha_0$ and $\alpha_n$ are constant terms related to the amplitude of time series, which are abbreviated as$(\alpha_0, \alpha_n)$. According to the sum-to-product identities of trigonometric functions, the above equation can be deformed to (6):

$$f(t) = \alpha_0 + \sum_{n=1}^{\infty} [\alpha_n \cos(n\omega t)\cos(\varphi) - \alpha_n \sin(n\omega t)\sin(\varphi)] \tag{6}$$

Then we multiply both sides of (6) by the function

$\sin(k\omega t)$ and integrate them over the period. Finally, the converting result can be denoted by (7).

$$\int_0^T f(t)\cdot\sin(k\omega t)\,dt = \int_0^T \alpha_0\cdot\sin(k\omega t)\,dt$$
$$+\int_0^T\sum_{n=1}^{\infty}\alpha_n\cdot\cos(n\omega t)\cdot\cos(\varphi)\cdot\sin(k\omega t)\,dt$$
$$+\int_0^T\sum_{n=1}^{\infty}\alpha_n\cdot\sin(n\omega t)\;\cdot\sin(\varphi)\cdot\sin(k\omega t)\,dt \qquad (7)$$

We defined an equation set $(\eta_n,\lambda_n)$: $\begin{cases}\eta_n=\alpha_n\cdot\cos(\varphi)\\ \lambda_n=-\alpha_n\cdot\sin(\varphi)\end{cases}$ as the set of amplitude variables that adapt to the form of (6). Then (6) can be converted into the following form:

$$f(t)=\alpha_0+\sum_{n=1}^{\infty}[\eta_n\,cos(n\omega t)+\lambda_n\,sin(n\omega t)] \qquad (8)$$

Furthermore, this formula needs to satisfy the constraint that the integrals of trigonometric functions of different frequencies over a period are zero, i.e.

$$\text{s.t.}\begin{cases}\displaystyle\int_0^T g(n\omega t)\,dt=0\\ \displaystyle\int_0^T g(n\omega t)\cdot h(m\omega t)\,dt=0\end{cases} \qquad (9)$$

$$\begin{cases}\eta_n=\dfrac{2}{T}\displaystyle\int_0^T f(t)\cdot\cos(n\omega t)\,dt=0\\ \lambda_n=\dfrac{2}{T}\displaystyle\int_0^T f(t)\cdot\sin(n\omega t)\,dt=0\end{cases} \qquad (10)$$

where both $g(n\omega t)$ and $h(m\omega t)$ are trigonometric functions with different frequencies. Next, we can find the specific solution to the system of equations $(\eta_n,\lambda_n)$, as illustrated in (10). Since $\alpha_0$ is a constant term and its corresponding frequency is 0. With the introduction of Euler's formula, (8) can be simplified as $f(t)=\sum_{-\infty}^{\infty}\frac{(\eta_n-j\lambda_n)}{2}e^{jn\omega t}=\sum_{-\infty}^{\infty}\mathcal{M}_n e^{jn\omega t}$. Then, $\mathcal{M}_n$ can be obtained by multiplying $e^{-jn\omega t}$ and integrating equally on both sides of $f(t)$, i.e.

$$\mathcal{M}_n=\frac{1}{T}\int_0^T f(t)\;e^{-jn\omega t}dt \qquad (11)$$

Therefore, the amplitude of the frequency wave can gain by calculating $|\mathcal{M}_n|$, the detailed derivation is shown in (12). $|\mathcal{M}_n|=\left|\frac{1}{T}\int_0^T f(t)\;e^{-jn\omega t}dt\right|$

$$=\left|\frac{1}{T}\int_0^T f(t)\,[\cos(-n\omega t)+i\sin(-n\omega t)]dt\right|$$
$$=\left|\frac{1}{T}\left(\int_0^T f(t)\cdot\cos(n\omega t)dt-i\int_0^T f(t)\cdot\sin(n\omega t)\,dt\right)\right|$$
$$=\left|\frac{1}{T}\left(\frac{T}{2}(\eta_n-i\lambda_n)\right)\right|$$
$$=\frac{1}{2}\alpha_n \qquad (12)$$

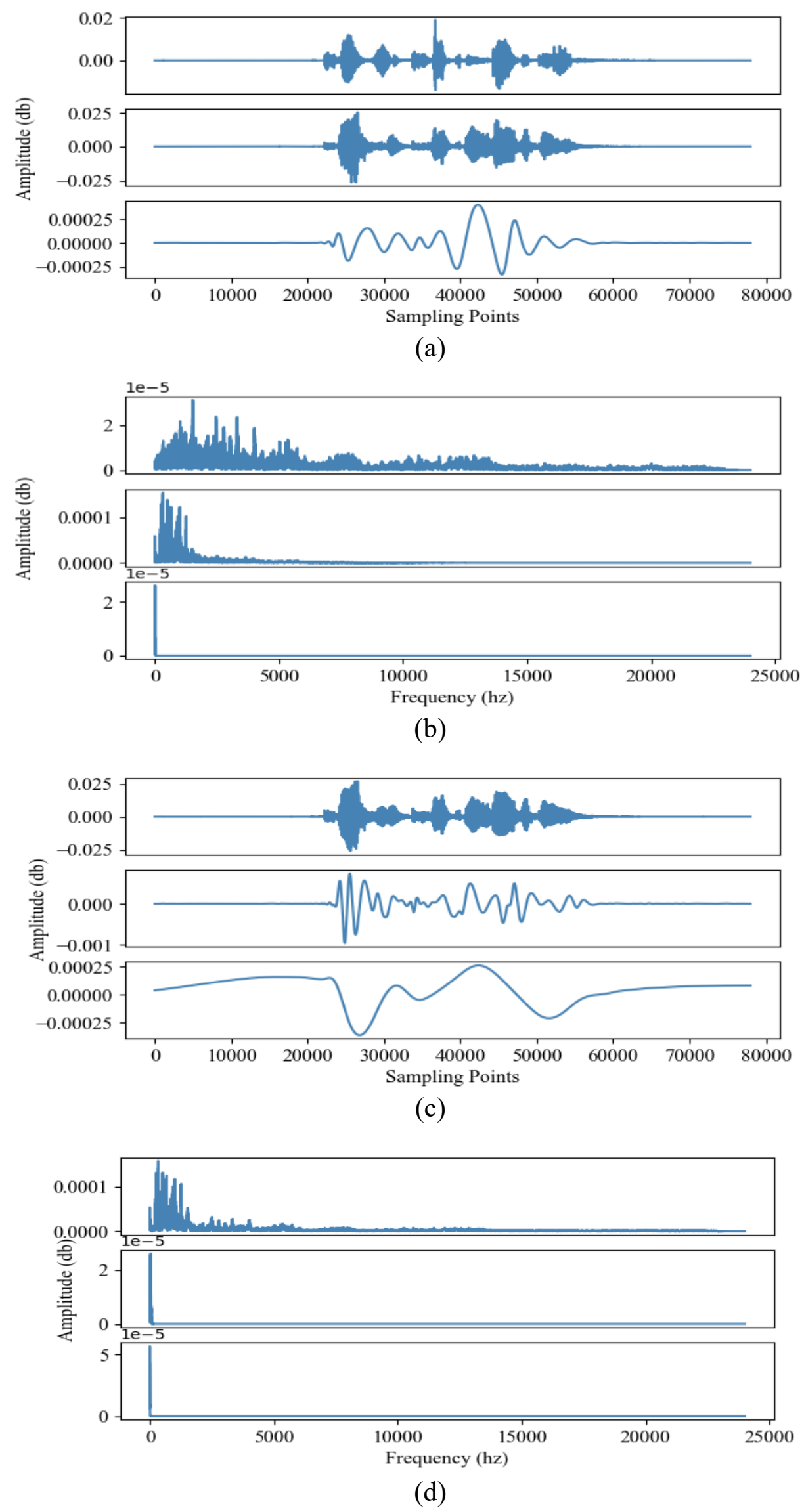

**Fig.4.** Comparison results of clustering method and grouping method. (a) and (b) show the time-domain characteristics and frequency-domain characteristics of the subsequence obtained by the clustering method, respectively. (c) and (d) represent the time-domain characteristics and frequency-domain characteristics of the subsequence obtained by the grouping method, respectively.

The variable $\mathcal{M}_n$ contains information about the frequency amplitude of the signal, i.e. $\alpha_n\to|\mathcal{M}_n|$, where '→' indicates the time-frequency mapping relation for the Fourier transforms. Then, amplitude conversion between the frequency domain and the spectrogram can be expressed as $\mathcal{A}_n=-20\cdot log(|\mathcal{M}_n|)$The formants are generally the maximum value of the approximate envelope of the frequency domain amplitude. Therefore, the relationship between the signal envelope and the formant can be expressed as $\alpha_n\leftrightarrow|\mathcal{M}_n|\leftrightarrow\mathcal{A}_n$. Moreover, due to the original signal can be reconstructed by superimposing the decomposed subsequences, thus the overall envelope information of the original signal can be reconstructed from the corresponding sub-series envelopes. Therefore, the formant information is expanded from the original single overall information to a local information set represented by multiple sub-sequences, which increases the fine-grained representation

of the formant information.

The variable $\mathcal{M}_n$ contains information about the frequency amplitude of the signal, i.e. $\alpha_n \rightarrow |\mathcal{M}_n|$, where '→' indicates the time-frequency mapping relation for the Fourier transforms. Then, amplitude conversion between the frequency domain and the spectrogram can be expressed as $\mathcal{A}_n = -20 \cdot log(|\mathcal{M}_n|)$. The formants are generally the maximum value of the approximate envelope of the frequency domain amplitude. Therefore, the relationship between the signal envelope and the formant can be expressed as $\alpha_n \leftrightarrow |\mathcal{M}_n| \leftrightarrow \mathcal{A}_n$. Moreover, due to the original signal can be reconstructed by superimposing the decomposed subsequences, thus the overall envelope information of the original signal can be reconstructed from the corresponding sub-series envelopes. Therefore, the formant information is expanded from the original single overall information to a local information set represented by multiple sub-sequences, which increases the fine-grained representation of the formant information.

### *D. Feasibility Analysis of Clustering Methods in AFMM*

To investigate the difference between the clustering method proposed in our method and the grouping method mentioned in [38], the same IMF sequence is used as the common input. Furthermore, the parameter settings of the EMD algorithm are set in line with [38] to maintain the consistency of the parameters. Therefore, we set the K=3 in the K-means algorithm according to the number of groupings described in [38]. The comparison results are shown in Fig.4.

From the comparison results shown in Fig. 3, we can see that the grouping method leads to information redundancy and information loss. The mid-frequency and low-frequency groups (as shown in the middle and bottom figures in Fig.4. (d)) are almost identical in the frequency domain, which results in information redundancy. Moreover, the grouping method aggregates the major temporal information in one subsequence (as shown in the top figure in Fig.4. (c)), so the rest of the subsequences (the bottom figure in Fig.4. (c)) are probably invalid, which might give rise to information loss. Thus, the grouping method breaks the sub-sequence completeness and reduces the sample diversity compared to the clustering method, which may affect experimental results.

## IV. Experiments

In this section, we conduct extensive experiments to evaluate the audio-based cross-modal auxiliary network on multimodal datasets.

### *A. Datasets*

The multimodal datasets RAVDESS [39], CMU-MOSI [40], and CMU-MOSEI [41] are selected to conduct our experiments. The dataset (RAVDESS) is employed to demonstrate the high efficiency of our model in the multimodal sentiment analysis task. All conditions are available in three modality formats: Audio-only (16bit, 48kHz .wav), Audio-Video (720p H.264, AAC 48kHz, .mp4), and Video-only (no sound). They are different representations of the same sentiment label, and they are synchronized in time**.** Moreover, the CMU-MOSI and CMU-MOSEI dataset is used to verify the advantage of the proposed algorithm by comparing it with other advanced methods. Both the video and audio files provided by CMU-MOSI and CMU-MOSEI are different modality formats extracted from the same audio-video sequence. The details about the training, testing, and validating split are presented in Table I, where the numbers in the table indicate the number of files. For example, '719' represents that RAVDESS contains 719 audio clip files.

TABLE I
Data clips and splits in the RAVDESS and CMU-MOSI datasets
(unit: number of files)

| Dataset | Audio | Video | Frames | Train | Valid |
|---|---|---|---|---|---|
| RAVDESS | 719 | 719 | 14125 | 11508 | 2617 |
| CMU-MOSI | 2199 | 2199 | 14574 | 10202 | 4372 |
| CMU-MOSEI | 3838 | 3837 | 18857 | 13200 | 5657 |

### *B. Baseline Models*

We conducted comparative experiments on several state-of-the-art multimodal fusion frameworks in two tests:

**Experiment-A**: We employ the CNN-based frameworks that include Alex-Net [42], Google-Net [43], Shuffle-Net [44], and Res-Net [45] as classifiers and denote them as VCAN-A, VCAN-G, VCAN-S, and VCAN-R, respectively. All mentioned neural networks use the standard network structure and hyper-parameter settings provided in the relevant papers. The details of this work are shown in *Section F*.

**Experiment-B***:* In this experiment, the advantages of VCAN in audiovisual sentiment analysis are illustrated by comparing it with different information fusion strategies or different network structure designs. The comparison methods include *LFN* [46], *SWAFN* [47], *BBFN* [48], *BC_LSTM* [49], *MMMU-BA* [50] and *Dialogue-RNN* [51]. The details of this work are shown in *Section G*.

**Environment Configuration:** All the algorithms are coded using python in PyCharm Community Edition 2020. For each algorithm configuration and each instance, we carry out five independent replications on the same AMD Ryzen 5 3500X 6-Core Processor CPU @ 3.60 GHz with 16.00-GB RAM and NVIDIA GeForce GTX 1660 SUPER GPU in the 64-bit Windows 7 professional Operation System.

**Model-related Parameters:** Based on the results of the similarity analysis of the subsequences, we artificially set the decomposition stop threshold, i.e., the number of decomposing layers is set to 12. The acoustic features extracted from AFMM are composed of rhythmic features and sound quality features. The rhythmic features include logarithmic energy values and their first- and second-order differences, short-time energy and its first- and second-order differences, sound pressure levels and their first- and second-order differences, and over-zero rates. The sound quality features contain absolute and relative frequency perturbations as well as absolute and relative amplitude perturbations. Specifically, the key parameters for generating the MEL spectrogram are the sampling rate (default value: 44100), the length of framing (default value: 512 (32ms)), the length of Hamming window (default value: 512 (32ms)), and the number of MEL filters (default value: 26). The structural parameters of the individual CNN networks in the collaborative classifier groups are consistent with the configuration elaborated in the related published paper. Moreover, we set dropout=0.5 as a measure of regularization. The ReLu and softmax activation functions are employed in the dense layers, and the final classification layer, respectively. Each network is trained for 100 epochs with batch size=32, and the Adam optimizer with cross-entropy loss function is utilized for optimization.

TABLE II
THE EVALUATION METRICS OF BASELINE MODELS IN THE RELATED WORK

| *Ref.* | Evaluation metrics | | | | | |
|---|---|---|---|---|---|---|
| | ACC -7 | ACC -5 | ACC -2 | F1 score | CORR | MAE |
| BC_LSTM[49] | -- | -- | √ | -- | -- | -- |
| MMMUBA[40] | -- | -- | √ | -- | -- | -- |
| DialogueRNN[51] | -- | -- | √ | √ | √ | √ |
| LFN[46] | √ | -- | √ | √ | √ | √ |
| SWAFN[47] | √ | -- | √ | √ | √ | √ |
| BBFN[48] | √ | -- | √ | √ | √ | √ |
| MISA[55] | √ | -- | √ | √ | √ | √ |
| MMIM[54] | √ | √ | √ | √ | √ | √ |
| VCAN(OUR) | √ | √ | √ | √ | √ | √ |

TABLE III
EVALUATION METRICS IN THE RAVDESS AND CMU-MOSI DATASETS

| Evaluation Metrics | | | | | |
|---|---|---|---|---|---|
| Accuracy | | Polarity metrics | F1_score | CORR | MAE |
| ACC-7 | ACC-5 | ACC-2 | | | |

## *C. Evaluation metrics*

As multimodal sentiment analysis tasks are transformed into image classification, we adopt different evaluation metrics for the above datasets. Table II illustrates the evaluation metrics of baseline models in the related work. Considering the sentiment polarities and the emotional categories in multimodal sentiment analysis, several metrics including ACC-7, ACC-5, and ACC-2 are used to evaluate the results of multi-category classification and polarity analysis, respectively [46]. Furthermore, the Mean Absolute Error (MAE) and F1 score are employed to test the model performance and the Correlation Coefficient (CORR) is leveraged to measure correlation [47]. The details about the evaluation metrics for different datasets in this paper are shown in Table III.

## *D. Parametric experiments and analysis*

We investigate some essential parameters in AFMM such as the number of decomposition layers and the choice of cluster algorithms to justify the parameters set.

**Decomposition Parameters Study**: EMD is an unsupervised decomposition method where the number of subsequences is related to the frequency characteristics of the input signal. Since the value of decomposing layers is artificially set to extend the diversity of subsequences, thus we find the optimal value of decomposing layers by measuring the similarity between subsequences using the Cosine Similarity (CS). The calculation results are shown in Table IV, where $\mathcal{L}$ denotes the number of signal decomposition layers and CS is the calculated value of Cosine Similarity. From the calculation results in Table IV, the optimal number of decomposing layers for EMD is 12, after which the similarity between the data tends to stabilize. So, the number of decomposing layers is set to 12.

TABLE IV
COSINE SIMILARITY BETWEEN SPEECH SUBSEQUENCES AT DIFFERENT DECOMPOSITION LEVELS

| $\mathcal{L}$ | 4 | 5 | 6 | 7 | 8 |
|---|---|---|---|---|---|
| CS | 0.55947 | 0.55939 | 0.55942 | 0.55951 | 0.55953 |
| $\mathcal{L}$ | 9 | 10 | 11 | **12** | 13 |
| CS | 0.55952 | 0.55952 | 0.55954 | **0.55955** | 0.55955 |
| $\mathcal{L}$ | 14 | 15 | 16 | 17 | 18 |
| CS | 0.55955 | 0.55955 | 0.55955 | 0.55955 | 0.55955 |

TABLE V
THE DISTRIBUTION (NUMBER) OF SENTIMENT DATA OBTAINED BY DIFFERENT VIDEO KEYFRAMING METHODS (UNIT: NUMBER OF FRAMES)

| Type | CLU | MC | RW | CMSM |
|---|---|---|---|---|
| CALM | 222 | 240 | 240 | 1914 |
| HAPPY | 221 | 240 | 240 | 1897 |
| SAD | 218 | 240 | 240 | 1949 |
| ANGRY | 220 | 240 | 240 | 1909 |
| FEARFUL | 219 | 240 | 240 | 1971 |
| DISGUST | 230 | 240 | 240 | 2040 |
| SURPRISE | 226 | 240 | 240 | 1838 |

TABLE VI
COMPARISON RESULTS OF DIFFERENT VIDEO KEYFRAME SELECTION METHODS

| Methods | CLU | MC | RW | CMSM |
|---|---|---|---|---|
| ACC-7 (%) | 63 | 70 | 70 | **79** |
| ACC-5 (%) | 75 | 77 | 76 | **85** |
| ACC-2 (%) | 14 | 33 | 33 | **55** |
| CORR | 0.1 | 0.44 | 0.5 | **0.5** |
| LOSS | 2.08 | 1.74 | 1.43 | **0.59** |
| MAE | 2.27 | 1.7 | 1.4 | **1.1** |

**Video Keyframes Selection**: It is important to choose the appropriate video emotional keyframes due to decision results are generated by the frames-based classifier groups. We compared the CMSM-based method with several novel video key-frame selection methods (including CLS [52] and MC, RW [53]) on the RAVDESS data. In this comparison experiment, the input images derive from the video clips labeled as 'HAPPY' and the Alex-Net is employed as the representative classifier. Table V illustrates the distribution of video keyframes captured by different methods on the same dataset, where the values in the table are the frame number of corresponding categories. The comparison results in Table VI demonstrate that the CMSM's metrics such as multi-class classification (ACC-7: 0.79 and ACC-5: 0.85) and sentiment polarity (ACC-2: 0.55) are optimal values compared to other comparative methods. Moreover, the errors including MAE: 1.1 and LOSS: 0.59 are the lowest in the comparison experiment. Therefore, the CMSM-based method outperforms other comparative algorithms in video keyframes selection.

**Feasibility Study of Auxiliary Mechanisms:** To investigate how auxiliary modal expands the multi-scale visual representation of acoustic features, we conduct a case study on an audiovisual auxiliary instance. The Schematic diagram of the case study is shown in Fig.5, where the number in the figure indicates the index of the video frame. As shown in Fig.5, we calculated the scores of each IMF and selected the top 10 video keyframes. The optimal set of frames (Mutual) is obtained by re-ranking the scores in the IMF-based framesets. The results are shown in Table VII, where the number in the figure indicates the index of the video frame. The observations presented in Table VII indicate that the acoustic auxiliary visual mechanism can significantly increase the probability of selecting implicit frames. For example, the IMF-0 provides other key-frame candidates (i.e. 11 and 17) in addition to the video key-frame indexes contained in the Mutual component. Thus, the auxiliary mechanisms can extract more comprehensive, refined, and efficient representations of emotion.

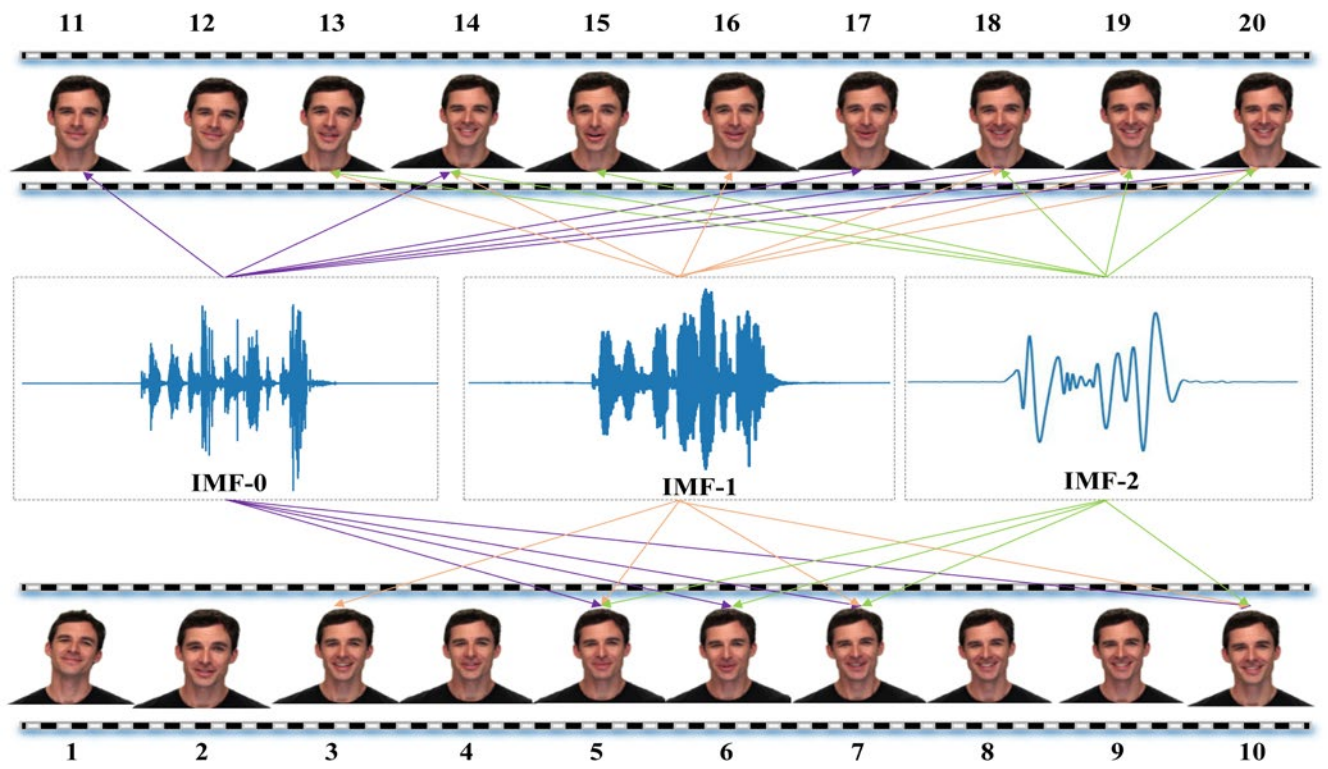

**Fig.5.** Schematic diagram of the case study

TABLE VII
KEYFRAMES SELECTED BY THE AUXILIARY MECHANISM IN DIFFERENT SUBSEQUENCES (IN ORDER)

| Component | No. |
|---|---|
| IMF-0 | 11,14,17,18,19,20,10,7,6,5 |
| IMF-1 | 13,14,16,18,19,20,3,5,7,10 |
| IMF-2 | 13,14,15,18,19,20,10,7,6,5 |
| Mutual | 14,18,19,20,5,7,10,13,6,16 |

### *E. Ablation Study*

To examine the function of the components in our model, we execute an ablation study on the RAVDESS [39] dataset. We divide experiments into two groups: spectrogram-based experiments and frame-based experiments. For spectrogram-based studies, the audio signals are transformed into the MEL spectrograms. Then, we use all spectrograms (Model-1), randomly selected spectrograms (Model-2), and AFMM-based spectrograms as inputs, respectively. For frames-based studies, we directly extract all frames, random frames, and CMSM-based frames from a video as the input of Model-3, Model-4, and CMSM, respectively. Consistent with the previous experiments, the Alex-Net is utilized as the classifier. The results are shown in Table VIII and Table IX, where NU and ME denote the number of extracted images and the memory occupied, respectively. From the Tables, we can observe that the AFMM indeed extends the diversity of audio representation (shown in Table VIII-NU/ME). However, the classification performance based on the AFMM's spectrogram is slightly inferior to other methods. The reasons are described in *Section H*. Comparing CMSM and Model_4, the overall performance of the former is better than that of the latter with almost the same input data. Especially, ACC-7 improved by nearly 10% (shown in Table IX: ACC-7). Although the performance of the CMSM is slightly lower than that of Module-3, the size of the input data of the latter is close to six times that of the former (shown in Table VIII: NU/ME). Therefore, we can conclude that CMSM can reduce redundant calculations while ensuring advanced classification performance.

TABLE VIII
RESULTS OF ABLATION EXPERIMENTS BASED ON MEL SPECTROGRAMS

| Methods | Model-1 | Model-2 | AFMM |
|---|---|---|---|
| ACC-7 (%) | 73 | 74 | 68 |
| ACC-5 (%) | 85 | 81 | 81 |
| ACC-2 (%) | 58 | 51 | 42 |
| CORR | 74 | 45 | 0.47 |
| LOSS | 1.21 | 0.66 | 1.52 |
| MAE | 0.92 | 1.35 | 1.57 |
| NU/ME | 672/34K | 91/4.5K | 2022/90.5K |

TABLE IX
RESULTS OF ABLATION EXPERIMENTS BASED ON VIDEO FRAMES

| Methods | Model-3 | Model-4 | CMSM |
|---|---|---|---|
| ACC-7 (%) | 81 | 71 | 80 |
| ACC-5 (%) | 88 | 82 | 86 |
| ACC-2 (%) | 62 | 50 | 60 |
| CORR | 0.55 | 0.51 | 0.51 |
| LOSS | 0.57 | 1.44 | 0.59 |
| MAE | 0.95 | 1.35 | 1.10 |
| NU/ME | 75K/7906K | 15K/1394K | 13K/1324K |

### *F. Results on the RAVDESS dataset*

Experiment-A is composed of three parts: spectrogram-based part, image (facial frames)-based part, and joint embedding-based part. Notably, the input for the joint embedding-based experiment is generated by VCAN. The experimental results are presented in Table X-XII, from which we can observe the following conclusion:

Compared with the results of the spectrogram-based experiment, the result in Table XI shows that the classification accuracies in the image-based experiment increased by 4.8%, 32.6%, 40.9%, and 41.3%, respectively. The comparison results based on joint embeddings are shown in Table XII, where 'A' denotes the spectrogram-only results, 'V' denotes the video image-only results, and 'AV' represents the experimental results based on the joint audiovisual embedding. Furthermore, '↑' indicates performance improvement, '↓' represents performance decrease, and '--' means no change in performance.

As presented in Table XII, the emotion recognition results of the joint embedding-based experiment are generally more advanced than that of spectrogram-based and image-based experiments. From the columns of the table, the optimal metrics in the experiment are ACC-7: 99.3%, LOSS: 0.41, CORR: 0.99 and MAE: 0.01 while that in other test (i.e. spectrogram-based / image-based) are ACC-7: 58.4% / 99.3%, LOSS: 0.96 / 0.43, CORR: 0.74 / 0.99 and MAE: 0.9 / 0.01, respectively. From the rows of the table, the performances of CNN models equipped with VCAN (i.e. 'AV') are significantly improved in all evaluation metrics compared with models without VCAN (i.e. 'A' and 'V'). Thus, it can be concluded that VCAN can effectively extract a more comprehensive and available audiovisual representation (compared with spectrum-based test) and select a more streamlined and accurate video frame (compared with image-based test).

TABLE X
EVALUATIONS OF SPECTROGRAM-BASED EMOTION RECOGNITION ON RAVDESS DATASET

| Methods | ACC-7 (%) | LOSS | CORR | MAE |
|---|---|---|---|---|
| Alex-Net | 53.5 | 1.29 | 0.68 | 1.09 |
| Google-Net | 47.8 | 2.07 | 0.67 | 1.13 |
| Shuffle-Net | 58.4 | 1.08 | 0.71 | 1.00 |
| Res-Net | 56.3 | 0.96 | 0.74 | 0.90 |

TABLE XI
EVALUATIONS OF IMAGE-BASED EMOTION RECOGNITION ON RAVDESS DATASET

| Methods | ACC-7 (%) | LOSS | CORR | MAE |
|---|---|---|---|---|
| Alex-Net | 58.3 | 0.68 | 0.58 | 1.05 |
| Google-Net | 80.4 | 0.79 | 0.81 | 0.50 |
| Shuffle-Net | 99.3 | 0.43 | 0.99 | 0.02 |
| Res-Net | 97.6 | 0.45 | 0.97 | 0.06 |

### *G. Results on CMU-MOSI and CMU-MOSEI dataset*

In this section, we conduct two sets of experiments on the state-of-the-art models mentioned in Experiment-B. One is a comparison with Tri-modalities methods on the CMU-MOSI dataset to illustrate the efficiency of VCAN in the absence of language modality. The other one is an experiment with Bi-modalities on CMU-MOSI and CMU-MOSEI datasets, which aims to demonstrate the high accuracy in identical conditions. The results of both VCAN and other methods on the CMU-MOSI dataset are shown in Table XIII and Table XIV. And the results of the CMU-MOSEI dataset are represented in Table XV. Specifically, all the benchmark methods in Table XIII are re-implemented and the relevant running environment configuration is described in *III.B*. The symbol '*' indicates results in the corresponding line are excerpted from previous papers [49-51].

The optimal metrics of VCAN-based methods in Table XIII are ACC-7: 36.9%, ACC-2: 75.9%, CORR: 0.64, F1 score: 0.76 and MAE: 0.93 while that for tri-modalities based methods are ACC-7: 34.4%, ACC-2: 76.6%, CORR: 0.66, F1 source: 0.76 and MAE: 0.96. The polarity classification result and correlation metrics of our model are slightly lower than those of the tri-modal fusion approach. This is probably since the text modality provides richer sentiment information, which results in the joint representations being more comprehensive. But our model significantly outperforms other models under the same input conditions, as demonstrated by comparing the ACC-2 between VCAN-based models (best: 75.9%) and Bi-modalities models (best: 65.2%).

The baseline models for the CMU-MOSEI experiment include BC-LSTM, BBFN (AV), MMMU-BA (AV), and SWAFN (AV), which predict the sentiment polarities by fusing audiovisual features. Notably, some baseline models provide variants for different modal combinations. For example, BBFN (AV) is a bi-modal variant of BBFN whose inputs are audiovisual modalities. From the experimental results reported in Table XV. we evaluate the VCAN on the CMU-MOSEI dataset and obtain 69.4% (VCAN-A), 71.3% (VCAN-G),72.5% (VCAN-S), and 73.9% (VCAN-R) accuracies with the bi-modal inputs. Compared to the accuracies (i.e., BC-LSTM: 60.7%, BBFN (AV): 71.1%, and SWAFN (AV): 58%) of baseline models, the proposed approach attains better performance.

TABLE XII
EVALUATIONS OF VIDEO EMOTION RECOGNITION ON RAVDESS DATASET

| Method | | ACC-7 (%) | LOSS | CORR | MAE |
|---|---|---|---|---|---|
| VCAN-A | AV | **60.5** | **0.59** | **0.50** | **1.05** |
| | A | ↑ 7.0 | ↑ 0.7 | ↑ 0.18 | ↑ 0.04 |
| | V | ↑ 2.2 | ↑ 0.09 | ↑ 0.08 | -- |
| VCAN-G | AV | **81.8** | **0.70** | **0.79** | **0.51** |
| | A | ↑ 34 | ↑ 1.37 | ↑ 0.12 | ↑ 0.62 |
| | V | ↑ 1.4 | ↑ 0.09 | ↓ 0.02 | ↑ 0.01 |
| VCAN-S | AV | **99.3** | **0.43** | **0.99** | **0.01** |
| | A | ↑ 40.9 | ↑ 0.65 | ↑ 0.28 | ↑ 0.99 |
| | V | -- | -- | -- | ↑ 0.01 |
| VCAN-R | AV | **98.7** | **0.41** | **0.99** | **0.03** |
| | A | ↑ 34 | ↑ 0.55 | ↑ 0.25 | ↑ 0.87 |
| | V | ↑1.4 | ↑ 0.04 | ↑ 0.02 | ↑ 0.03 |

TABLE XIII
EXPERIMENTAL RESULTS WITH THE BENCHMARK MODELS (TRI-MODALITIES)

| Methods | ACC-7 | ACC-2 | CORR | F1 | MAE |
|---|---|---|---|---|---|
| BBFN [48] | 25.6 | 68.8 | 0.50 | 0.69 | 1.20 |
| LFN [46] | 29.5 | 69.9 | 0.54 | 0.7 | 1.22 |
| SWAFN [47] | 34.4 | **76.6** | **0.66** | **0.76** | 0.96 |
| VCAN-A | 32.1 | 74.3 | 0.59 | 0.74 | 1.01 |
| VCAN-G | 34.2 | 74.2 | 0.61 | 0.74 | 0.99 |
| VCAN-S | 35.4 | 75.8 | 0.63 | 0.76 | 0.95 |
| VCAN-R | **36.9** | **75.9** | **0.64** | **0.76** | **0.93** |

TABLE XIV
EXPERIMENTAL RESULTS WITH THE BENCHMARK MODELS (BI-MODALITIES)

| Methods | ACC-2(%) | Methods | ACC-2(%) |
|---|---|---|---|
| VCAN-A | 74.3 | BC-LSTM [49]* | 62.1 |
| VCAN-G | 74.2 | MMMU-BA [50]* | 65.2 |
| VCAN-S | 75.8 | Dialogue-RNN [51]* | 63.2 |
| VCAN-R | **75.9** | -- | -- |

TABLE XV
EXPERIMENTAL RESULTS WITH THE BENCHMARK MODELS

| Methods | ACC-2 (%) | Methods | ACC-2 (%) |
|---|---|---|---|
| VCAN-A | **69.4** | BC-LSTM [49]* | 60.7 |
| VCAN-G | **71.3** | BBFN(AV) [48]* | 71.1 |
| VCAN-S | **72.5** | SWAFN(AV) [47]* | 58.0 |
| VCAN-R | **73.9** | MMMU-BA(AV) [50]* | **76.6** |

TABLE XVI
RECOGNITION RESULTS OF MMIM AND MISA IN CMU-MOSI DATASET

| Methods | ACC-7 | ACC-5 | ACC-2 | CORR | F1 | MAE |
|---|---|---|---|---|---|---|
| MMIM | 45.6 | 51.4 | 83.3 | 0.78 | 0.83 | 0.73 |
| MISA | 39.6 | -- | 79.4 | 0.74 | 0.81 | 0.83 |

### *H. Analysis and Discussion*

**The results analysis of Experiment-A:** The unimodal that the facial images are more efficient than spectrograms for emotion recognition. This is probably due to the inconsistent properties of the pixels. Feature extraction of images focuses on considering the spatial or temporal association of local pixel blocks (rather than pixel points), which facilitates the acquisition of fine-grained features for the sentiment. However, in the spectrogram, each pixel has an actual physical significance, i.e. the frequency and intensity of audio at a given moment. The process of convolution and pooling of local pixels disrupts the time-frequency distribution relationship and thus introduces more error terms. So, the facial images are more suitable than spectrograms for the CNN-based framework. The VCAN is more effective than unimodal networks. The reasons for this can be summarized in two aspects. Firstly, our model extends the emotional representation of audio modality and eliminates the uncertainty impact on the classification model caused by redundant frames. Secondly, VCAN identifies representative emotional pictures with the assistance of speech modalities by computing emotional-related statistical components in spectrograms. This strategy enhances multimodal information interaction by mimicking human emotional perception processes.

**The results analysis of Experiment-B**: We can observe that some tri-modal algorithms are not as effective as the VCAN from Table XIII. This is probably because the network structure disrupts the invariance of the modal information. For example, the specific component such as the modality-shared encoding matrix ignores the independence of each modal and then leads to information confusion and redundancy between modalities when fusing different modal features. On the contrary, VCAN can maintain modal interaction while ensuring information invariance of single modalities. Moreover, the audio decomposition increases the multi-scale affective representation and the feature clustering enhances the joint representation of similar acoustic vectors. Therefore, our model can achieve advanced results.

Additionally, the tri-modal fusion strategies obtain advanced results in some evaluation metrics as the textual modality plays an irreplaceable role in multimodal sentiment analysis. For example, the sentiment polarity classification results (ACC-2) of SWAFN outperform that of VCAN. This is probably because the language modality provides high-dimensional, comprehensive, and contextually relevant semantic information, which is important for enhancing the information completeness of the representation. In addition, SWAFN implements visual-textual fusion and auditory-textual fusion using textual modality as an auxiliary modality, which is structurally similar to VCAN. Combining the advantages of the textual modality and auxiliary mechanism, SWAFN obtains advanced classification results.

Notably, the accuracy of MMMU-BA is 76.6% in Table XV, which is better than that of VCAN. We believe that the attention mechanism in MMMU-BA plays a decisive role. Compared to VCAN, the attention mechanism can mine deeper joint sentiment representations with automatic semantic alignment. However, combined with the performance of MMMU-BA on the CMU-MOSI dataset, the attention mechanism is easily affected by the size of the data. The attention mechanism model is more inclined to train model parameters on large-scale data. In addition, the parallel encoding and independent decoding structure of the attention mechanism can aggravate the time and memory consumption of model training. By comparing the error span (CMU-MOSI Accuracy: MMMU-BA-65.2% vs VCAN:75.9% and CMU-MOSEI Accuracy: MMMU-BA-76.6 % vs VCAN:73.9%) of VCAN and MMMU-BA, we can observe that the overall performance of VCAN is better than that of MMMU-BA.

**Discussion**: Although the auxiliary mechanism gets advanced results in multimodal sentiment analysis, both the transition between feature vectors and special fusion strategies such as the treatment of imbalanced contribution for different modalities are indispensable. For example, both the MMIM [54] and MISA [55] apply specialized feature extractors to create representative vectors that can model temporal dependencies of input sequences. Moreover, the innovative data structure transformations are utilized in the transitional stages, e.g. the Modality Invariant Encoder module and the Contrastive Predictive Coding. The classification results are shown in Table XVI, which exhibits a high-level performance in emotion recognition. Similarly, the SWAFN explores the imbalanced contribution of the different modalities and gives a reasonable solution. These issues are not considered in VCAN, and we will design an improved network that combines them in future works.

Furthermore, we briefly analyze the feasibility of VCAN combined with language modality. Language modality has rich semantic information, but we find that embedding language modalities into VCAN may degrade the classification performance. We list several key technical bottlenecks that prevent the embedding of language modalities into VCAN. The specific analysis is as follows: 1. Since language modality has rich semantic information, it plays an important role in multimodal sentiment analysis. However, language modalities have the potential to disrupt modal consistency and cause semantic confusion. Take Fig.6 as an example to illustrate this problem. As shown in Fig.6. (A), the emotional performance of the actor is “depressed”, which is marked by the red solid line box in the diagram. But the emotional keyword conveyed by the text in the picture is “Happy”, which is indicated by the red dashed line box. The semantics of the two modalities are opposite. When incorporating textual semantic information into audio-visual modalities, the decision network may output incorrect classification due to inconsistent emotional labels of the identical frame. Therefore, language modalities may interfere with the overall decision when multimodal affective representations are inconsistent. 2. VCAN relies on the semantic and temporal synchronization of audiovisual modalities, i.e., audiovisual emotional representations are consistent at the same moment. However, since the textual information of multiple consecutive frames in a video is almost constant, a large number of redundant or mismatched video keyframes may be introduced if language modalities are embedded in VCAN. This problem can be seen in Fig.6.(B). The last three frames provide the main contribution to determining the character's emotions, which can be selected by VCAN. However, if language modalities are incorporated into the cross-modal selection mechanism, each frame in Fig.6. (B) has the same weight, which will increase the risk of false selection. 3. Compared to visual and audio modalities, language modality is capable of generating high-dimensional semantic vectors, which may result in imbalance contribution during the multimodal fusion. In other words, the network training process may excessively boost the contributions of text modalities while ignoring the contributions of other modalities. A detailed analysis of the issue can be found in [56].

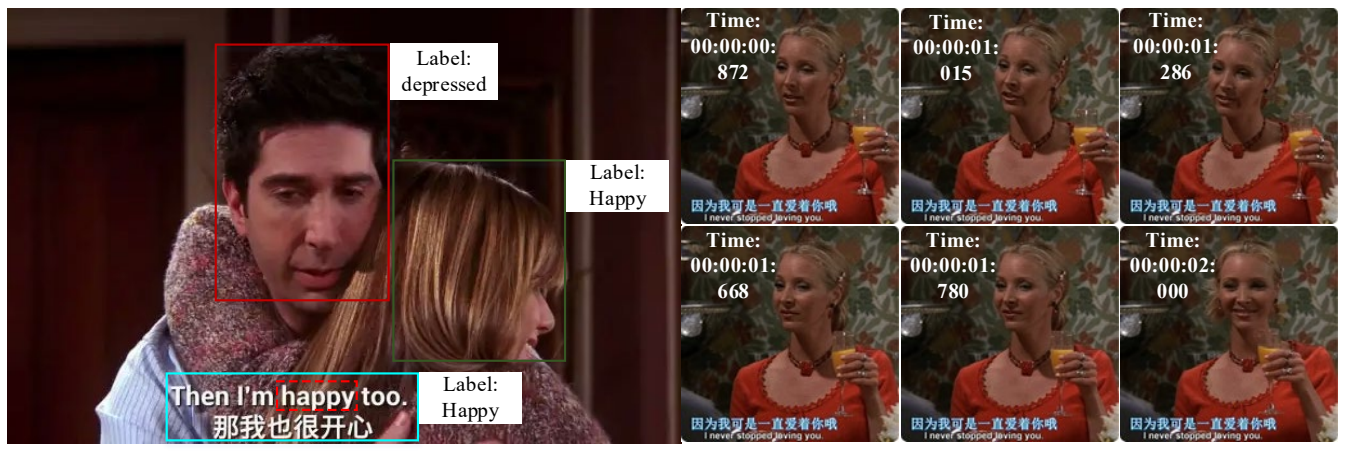

(A) (B)

**Fig.6.** The problems of VCAN combined with language modality.

## V. Conclusion

A video-based cross-modal auxiliary network (VCAN) for multimodal sentiment analysis has been proposed to enhance acoustic feature diversity before multimodal fusion and reduce feature redundancy in audiovisual interactions. In VCAN, multi-level acoustic representations are generated based on the audio feature mapping module, which increases the emotional representations of acoustic modality in different frequency bands. Furthermore, the cross-modal selection module is proposed to efficiently filter redundant frames for audiovisual feature fusion. Six advanced algorithms (including LFN, SWAFN, BBFN, BC_LSTM, MMMU-BA, and Dialogue-RNN) are introduced as benchmark methods. Extensive experimental results on RAVDESS and CMU-MOSI datasets indicate that VCAN is significantly superior to these state-of-the-art methods. In the future, we will further investigate the imbalance of modal contributions in multimodal sentiment analysis through mutual information methods and subspace theory.

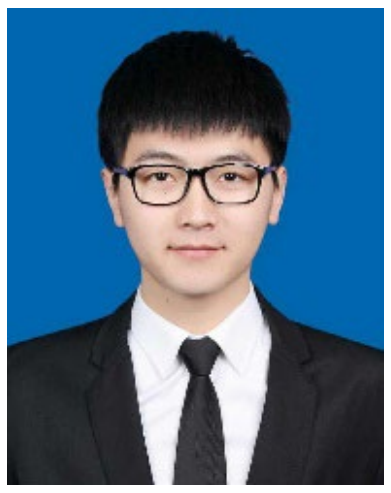

**Rongfei Chen** received his B.S. degree from Ludong University in 2017 and his M.S. degree from Yanshan University in 2020, respectively. He is currently a Ph.D. candidate in Control Science and Engineering at Shanghai University. His research interests include multimodal learning and sentiment analysis

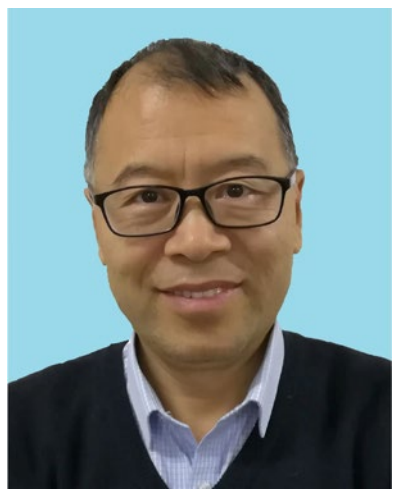

**Wenju Zhou** received his B. Sc. degree and M. Sc. degree both from Shandong Normal University in 1990 and 2005 respectively and obtained a Ph.D. degree from Shanghai University in 2014. He is now a professor at Shanghai University. His research interests include medical image processing, machine vision, and the industrial applications of automation equipment.

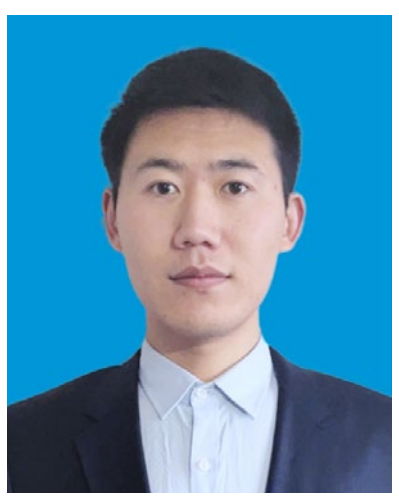

**Yang Li** received his B.S. degree in 2016 and M.S. degree in 2019, both from Qingdao Agricultural University and Shandong University of science and technology, China. And currently he is a Ph.D. candidate in Control Science and Engineering in Shanghai University. His research interests are in artificial intelligence and machine learning.

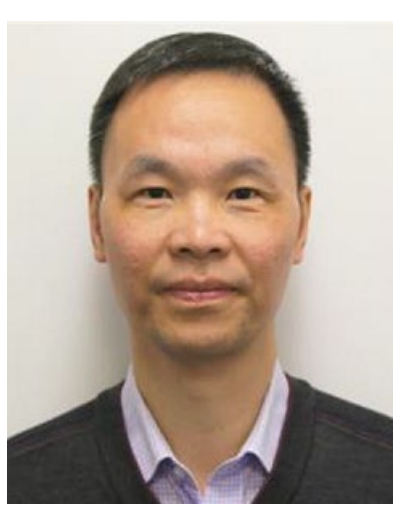

**Huiyu Zhou** received his B.E. degree in radio technology from Huazhong University of Science and Technology, China, the M.Sc. degree in biomedical engineering from University of Dundee, U.K., and the D.Phil. degree in computer vision from Heriot-Watt University, Edinburgh, U.K. He is currently a Professor with School of Computing and Mathematical Sciences, University of Leicester, United Kingdom.